# HS-SLAM: A Fast and Hybrid Strategy-Based SLAM Approach for Low-Speed Autonomous Driving

Bingxiang Kang[1], Jie Zou[1], Guofa Li[1,*], Pengwei Zhang[2], Jie Zeng[3], Kan Wang[3], Jie Li[1]

[1]CQU  [2]BYD  [3]CMTT

**Abstract**
Visual-inertial simultaneous localization and mapping (SLAM) is a key module of robotics and low-speed autonomous vehicles, which is usually limited by the high computation burden for practical applications. To this end, an innovative strategy-based hybrid framework HS-SLAM is proposed to integrate the advantages of direct and feature-based methods for fast computation without decreasing the performance. It first estimates the relative positions of consecutive frames using IMU pose estimation within the tracking thread. Then, it refines these estimates through a multi-layer direct method, which progressively corrects the relative pose from coarse to fine, ultimately achieving accurate corner-based feature matching. This approach serves as an alternative to the conventional constant-velocity tracking model. By selectively bypassing descriptor extraction for non-critical frames, HS-SLAM significantly improves the tracking speed. Experimental evaluations on the EuRoC MAV dataset demonstrate that HS-SLAM achieves higher localization accuracies than ORB-SLAM3 while improving the average tracking efficiency by 15%.

## 1 Introduction

Simultaneous localization and mapping (SLAM) is a key technology for robotics and low-speed autonomous vehicles [1-3]. As shown in Fig. 1, existing SLAM algorithms can be broadly classified into three categories based on their methodology and sensor modalities: direct SLAM, feature-based SLAM, and visual-inertial SLAM. Among these categories, visual SLAM has gained significant attention for its low-cost deployment, rich perceptual information, and structural simplicity. However, visual SLAM suffers from scale ambiguity, sensitivity to dynamic environments, and motion blur, which can lead to degraded localization accuracy and tracking failures in challenging scenarios. Feature-based SLAM is prone to failure in textureless or repetitive environments due to the lack of distinct feature points, while direct SLAM is more susceptible to photometric inconsistencies and requires a well-calibrated camera. Therefore, visual-inertial SLAM technologies have been developed for improvement due to the complementary nature of cameras and inertial sensors [4-7].

Visual-inertial SLAM has emerged as a robust and effective solution by fusing visual data with inertial measurements to resolve scale ambiguity and enhance state estimation. Early approaches leverage inertial measurements during initialization to recover metric scale, gravity direction, velocity, and sensor biases, typically by first performing visual structure-from-motion (SfM) and subsequently aligning it with Inertial Measurement Unit (IMU) preintegration within a sliding window [8]–[9]. While stereo systems directly compute depth to overcome scale ambiguity, they often incur higher computational and hardware costs [10]. Filter-based visual-inertial SLAM methods, primarily derived from the Multi-State Constraint Kalman Filter (MSCKF) framework, alternate between IMU propagation and visual updates to reduce state dimensionality. However, issues such as linearization errors and observability mismatches have driven enhancements including first-estimates Jacobian techniques, explicit observability constraints, and robot-centric formulations [11-13]. In contrast, optimization-based methods, exemplified by systems such as VINS-mono [14], ORB-SLAM3 [15], and DM-VIO [16], employ sliding-window bundle adjustment to improve accuracy despite challenges in maintaining consistency and real-time performance.

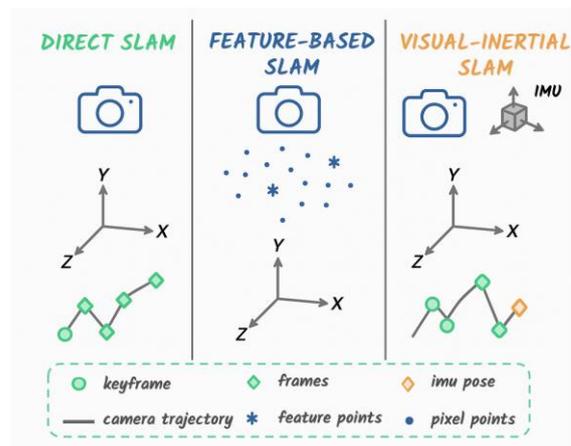

**Fig. 1** Overview of Direct, Feature-Based, and Visual-Inertial SLAM

Furthermore, learning-based approaches have advanced visual-inertial SLAM by enhancing feature extraction and multimodal data fusion through geometric networks and attention mechanisms, thereby increasing robustness in dynamic environments [17-18].

---
* Corresponding author.

However, for practical applications, these visual-inertial SLAM systems are required to be feasible on small and resource-limited embedded devices. To this end, this paper aims to propose a fast visual-inertial SLAM approach by focusing on enhancing the localization efficiency. The main contributions are summarized as follows:

1) We propose a novel SLAM framework that strategically integrates feature-based and direct methods. Experimental results demonstrate that HS-SLAM surpasses ORB-SLAM3 in both localization accuracy and computational efficiency.
2) We develop a four-stage optimization pipeline consisting of map initialization, current frame tracking, local map tracking, and keyframe selection. This pipeline minimizes unnecessary feature extraction and descriptor computation in non-keyframes, thereby significantly improving operational efficiency.
3) A two-stage fusion framework is introduced for front-end visual odometry tasks. In the first stage, IMU prior information is employed for robust motion prediction, while in the second stage, a multi-layer image pyramid architecture is utilized to progressively refine precision.

## 2 Related Work

### 2.1 Feature-Based Visual SLAM

Feature-based visual SLAM generally relies on feature extraction, feature description, and feature matching to accomplish simultaneous localization and mapping tasks. Davison et al. proposed MonoSLAM [19], the first real-time monocular visual SLAM framework, which employs a random search mechanism in the front-end to convert captured 3D object information into feature representations in the 2D image space. Raul Mur-Artal et al. subsequently developed ORB-SLAM [20] and ORB-SLAM2 [10]. The ORB-SLAM family utilizes ORB feature points to balance computational efficiency and accuracy. Carlos et al. later introduced ORB-SLAM3 [15], which extended support for fisheye cameras. Since then, ORB-SLAM has supported monocular, stereo, and RGB-D cameras and introduced the concept of multi-map systems.

Yang et al. proposed UPLP-SLAM [21], which unifies point, line, and plane features to improve pose estimation accuracy and robustness in structured environments. Xiang Jia et al. further introduced EGLT-SLAM [22], incorporating entropy theory to enhance line feature extraction. LIFT-SLAM [23], proposed by Bruno et al., replaces traditional geometric methods with neural networks for feature extraction, achieving more accurate feature matching. Shao et al. [24] proposed a Faster R-CNN-based semantic filter to refine feature correspondences, improving epipolar geometry estimation and vSLAM localization accuracy. As illustrated in Fig. 2, recent advancements in feature-based SLAM have gradually integrated structural feature representations, deep learning-based feature extraction, semantic perception, and entropy modeling.

### 2.2 Direct Visual SLAM

The direct method assumes gray-level invariance in image pixels and estimates the intensity map transformation by minimizing pixel-wise gray-level discrepancies. Based on the density of pixel utilization, direct methods can be classified into dense and semi-dense approaches. Dense methods like DTAM [25] reconstruct depth maps using all available image pixels, achieving high reconstruction density at the cost of substantial computational overhead. In contrast, semi-dense methods such as LSD-SLAM [26], DSO [27], and SVO [28] focus on informative pixels, typically those with strong gradients or located at corners, to balance computational efficiency with mapping accuracy.

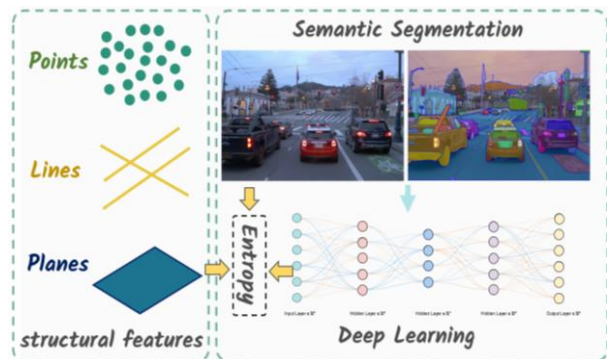

**Fig. 2** Overview of recent advancements in feature-based SLAM, illustrating the integration of geometric primitives and entropy-aware deep learning for enhanced semantic perception.

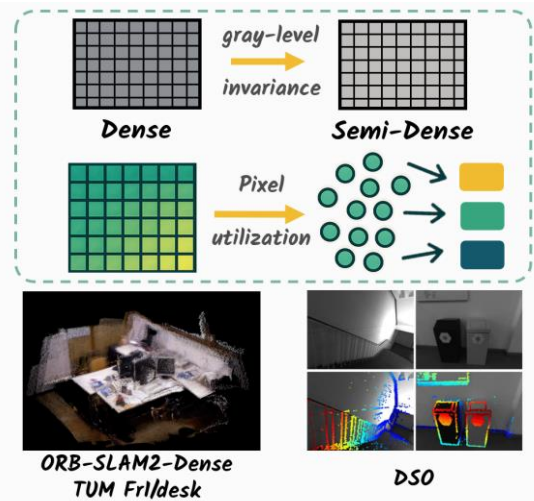

**Fig. 3** Gray-Level Invariance Assumption and Pixel Utilization in Dense and Semi-Dense Direct SLAM Methods [27].

Fig. 3 illustrates this distinction. The left image shows a dense reconstruction result generated using our extended ORB-SLAM2 system with a dense mapping module, evaluated on the TUM RGB-D dataset (Fr1/desk sequence). The right image is extracted from the original DSO paper [27], showcasing a semi-dense reconstruction where only a subset of high-gradient pixels contributes to the map. Notably, the variation in color intensity across the semi-dense map reflects the differing weights or confidence levels assigned to each feature point during reconstruction.

Compared to feature-based methods, direct methods eliminate the need for feature extraction, maintain stability in low-texture environments, and support semi-dense or dense map reconstruction. However, their performance is highly sensitive to lighting variations, which can compromise the assumption of gray-level invariance and potentially lead to tracking failure.

### 2.3 Visual–Inertial SLAM

Pure visual SLAM often struggles with localization in challenging scenarios, such as rapid motion or lens occlusion. In contrast, inertial measurement units (IMUs) can provide stable motion estimates even when the camera encounters issues such as feature loss, lighting variations, or severe jitter by measuring angular velocity and acceleration. Although IMUs suffer from zero-bias divergence, this issue can be mitigated through visual corrections. Consequently, integrating IMUs with visual SLAM to form visual-inertial SLAM significantly enhances system robustness and accuracy.

Among tightly coupled approaches, OKVIS [29], proposed by Leutenegger et al., is a classical method that formulates both vision and IMU residual terms and jointly optimizes state variables. VINS-Mono [14], developed by Qin et al., further refines this framework by enabling online sensor delay estimation, employing a sliding window to optimize back-end bundle adjustment (BA) computations, and leveraging optical flow for front-end feature tracking. Tsuei et al. [30] addressed systematic inaccuracies in covariance estimation for visual-inertial localization by learning a nonlinear mapping from empirical ground truth to EKF-estimated values, improving localization robustness. Yin et al. proposed Dynam-SLAM [31], a visual-inertial SLAM system that tightly integrates IMU measurements with feature observations to enhance robustness and accuracy in dynamic environments. Additionally, Ground-Fusion [32], proposed by Yin et al., combines RGB-D camera, IMU, wheel odometer, and GNSS sensor, specifically designed for corner cases in low-power scenarios.

However, while existing methods have demonstrated significant advancements, their computational complexity and resource demands often limit their deployment on small-scale devices. To overcome these challenges, this study proposes a novel strategy that integrates feature-based and direct methods within a visual-inertial SLAM framework, aiming to achieve efficient and high-performance localization.

## 3 System Description

The HS-SLAM system enhances efficiency through strategic minimization of non-essential descriptor extraction, thereby enabling optimal deployment on resource-constrained devices with limited computational power (e.g., embedded platforms, low-speed autonomous vehicles). An overview of the proposed system is shown in Fig. 4. Built upon ORB-SLAM3, HS-SLAM introduces an innovative fusion of feature-based and direct methods, integrating them into the front-end tracking module. The system processes stereo images, where Cam0 and Cam1 correspond to the left and right cameras, respectively, and incorporates IMU data as input, employing a hybrid feature-direct fusion strategy in the front-end.

The back-end relies on local mapping, loop map merging, and full bundle adjustment (BA), followed by a global atlas management module. The fusion of feature-based and direct methods allows for robust tracking and mapping, even in environments with limited resources. A detailed discussion of HS-SLAM's hybrid

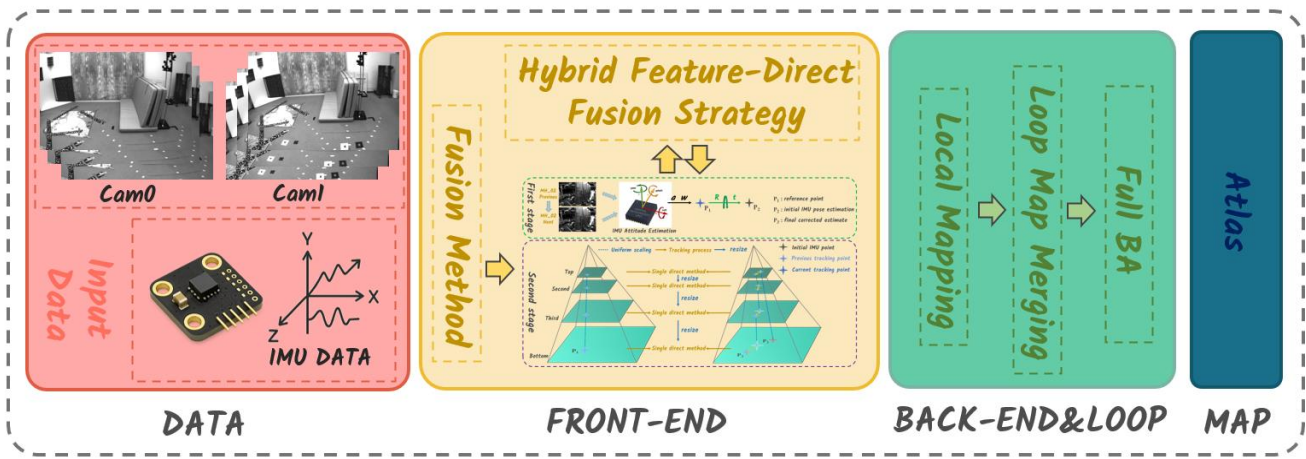

**Fig. 4** Overview of the HS-SLAM system

feature-direct fusion strategy and integration approach will be provided in subsequent sections.

## 3.1 Hybrid Feature-Direct Fusion Strategy

The fusion strategy flowchart consists of four main steps: map initialization, current frame tracking, local map tracking, and keyframe selection. To better illustrate the details of our method for improved understanding, the detailed algorithmic process is as follows. Given the complexity of variables and functions, Table 1 provides a comprehensive explanation of the functions and variables used in Algorithm 1. To keep the algorithm's pseudocode as concise as possible, we have adopted extensive abbreviations.

### 3.1.1 Map Initialization

The initialization process begins by extracting ORB feature points and corresponding descriptors from the first two frames, denoted as $I_1$ and $I_2$. These feature points, represented as $p_i^1$ and $p_i^2$, are matched based on their descriptors $d_i^1$ and $d_i^2$, forming a set of 2D-2D correspondences $\mathcal{M}$:

$$\mathcal{M} = \{(p_i^1, p_i^2) | d_i^1 \approx d_i^2\} \quad (1)$$

Given these correspondences, the essential matrix $E$ is estimated using epipolar geometry, where $E$ is derived as:

$$E = K^T F K \quad (2)$$

where $F$ is the fundamental matrix and $K$ denotes the camera intrinsic matrix. The camera pose $(R, t)$ is then recovered by decomposing $E$ using singular value decomposition (SVD):

$$E = U \Sigma V^T \quad (3)$$

where $U$ and $V$ are orthogonal matrices, and $\Sigma$ is a diagonal matrix containing the singular values of $E$. After selecting the physically valid solution, the initial camera pose is determined. With this pose established, the 3D map points $X_i$ are then computed using triangulation as:

$$X_i = Triangulate(p_i^1, p_i^2, R, t) \quad (4)$$

The set of triangulated points forms the initial map, denoted as $P = \{X_i\}$.

### 3.1.2 Current Frame Tracking

In this section, we describe the fusion method and 2D-3D reprojection optimization used to track the current frame. The 2D-3D reprojection optimization is performed to refine the pose estimation. The reprojection error for each point is given by:

$$e_i = \prod(T_t X_i) - p_i^t \quad (5)$$

where $T_t$ is the estimated transformation (pose) at time $t$. The reprojection error $e_i$ measures the difference between the projected 3D points and their corresponding 2D image locations. Here, $p_i^t$ represents the observed 2D pixel coordinates of the 3D point in the image frame at time $t$. The transformation $\hat{T}_t$ is then refined by minimizing the following objective function using nonlinear optimization (Gauss-Newton method):

$$\hat{T}_t = \arg \min_{T_t} \sum_i \|e_i\|^2 \quad (6)$$

If the reprojection error exceeds a predefined threshold for certain points, and the remaining valid matches are fewer than a minimum threshold $N_{min}$, the tracking is considered to have failed. When tracking fails,

---

**Algorithm 1: Hybrid Feature-Direct Fusion Strategy**

**Input:** $I_1, I_2, \{I_t\}$; $N_{min}$, threshold; $\tau_T, \tau_p$

**Output:** $P$; $T_t$
(features$_{1,2}$ desc$_{1,2}$) ← EO($I_1, I_2$); Matches ←MF(desc$_1$, desc$_2$)
$E$ ← EEM(Matches); $(R, t)$ ← DEM($E$)
$P$ ← TP(Matches, $R$, $t$); previousPose ← IdentityMatrix
**for** each new frame $I_t$ **do**
  FusionMatches ← FFM($I_t$, $P$); $T_t$← OP(FusionMatches, $P$)
  **if** EPE($T_t$, $P$) > threshold **or** VM(FusionMatches) < $N_{min}$ **then**
    $T_t$ ←RWK($T_t$)
    **if** ($T_t$ is invalid) **then** $T_t$ ← GR($I_t$) **end if**
  **end if**
  **if** DA ($I_t$) == false) **then**
    $\mathcal{M}_t$ ← OpticalFlow($I_{t-1}, I_t$)
  **else**
    $\mathcal{M}_t$← MF(KeyframeDescriptors, ED($I_t$)); Filter($\mathcal{M}_t$)
  **end if**
  poseChange ← FN(Inverse($T_t$) × previousPose − I)
  avgParallax ← AP($\mathcal{M}_t$)
  **if** (poseChange > $\tau_T$) **or** (avgParallax > $\tau_p$) **or** (Count(M$_t$) < $N_{min}$) **then**
    $KF_t$ ←$I_t$
    **if** (ORBFeatures not extracted) **then** (KFfeatures, KFdesc) ← EO($KF_t$) **end if**
    UMWK($KF_t$)
  **end if**
  previousPose ←$T_t$
**end for**

Table 1. Explanation for algorithm 1 variables and functions.

| Variable/Function | Explanation |
|---|---|
| $I_1, I_2, \{I_t\}$ | Two initial frames and Sequence of subsequent frames |
| $N_{min}$, threshold | Minimum valid matches and Reprojection error threshold |
| $\tau_T, \tau_p$ | Pose change threshold and Parallax threshold |
| $P, T_t$ | Set of 3D map points and Optimized camera poses for each frame |
| features$_{1,2}$ | Feature points from the first and second frames |
| desc$_1$, desc$_2$ | Descriptors of the corresponding feature points |
| Matches | Matched feature pairs between the two frames |
| EO() | ExtractORB() |
| MF() | MatchFeatures() |
| EEM() | EstimateEssentialMatrix() |
| DEM() | DecomposeEssentialMatrix() |
| TP() | TriangulatePoints() |
| FFM() | FusionFeatureMatching() |
| OP() | OptimizePose() |
| EPE() | eprojectionError() |
| VM() | ValidMatches() |
| RWK() | RelocalizeWithKeyframe() |
| OF() | OpticalFlow() |
| GR() | GlobalRelocalization() |
| DA() | DescriptorsAvailable() |
| ED() | ExtractDescriptors() |
| FN() | FrobeniusNorm() |
| AP() | AverageParallax() |
| UMWK() | UpdateMapWithKeyframe() |

the system first attempts to re-establish tracking using a reference keyframe $KF_r$ and the current frame $I_t$ through relocalization. This is done by solving:

$$T_t = Relocalization(KF_r, I_t) \quad (7)$$

If relocalization fails, a global relocalization attempt is made by minimizing the photometric error between the database image $I_{DB}$ and the current frame $I_t$:

$$T_t = arg\min_{T_t} \sum_i \|I_{DB}(p_i) - I_t(Tp_i)\|^2 \quad (8)$$

### 3.1.3 Local Map Tracking

When tracking the local map, the system first decides which method to use depending on whether descriptors have been extracted. If descriptors are not available, the optical flow method is employed. Optical flow computes the displacement of image points between consecutive frames, providing a set of correspondences. Given two consecutive images $I_{t-1}$ and $I_t$, the optical flow equation is expressed as:

$$p_i^t = p_i^{t-1} + \Delta p_i \quad (9)$$

where $p_i^{t-1}$ and $p_i^t$ are the positions of the point $i$ in the previous and current frames, respectively, and $\Delta p_i$ is the flow vector that corresponds to the motion of the point from the previous frame to the current frame. The set of correspondences between the frames is then denoted as:

$$\mathcal{M}_t = OpticalFlow(I_{t-1}, I_t) \quad (10)$$

where $\mathcal{M}_t$ represents the set of tracked feature points at time $t$. Alternatively, if descriptors have been extracted or if optical flow tracking fails, the system switches to feature point matching. In this case, each keypoint in the keyframe $KF$ is matched to a corresponding point in the current frame $I_t$ based on the similarity of their descriptors. The feature matching criterion is formally expressed as follows:

$$\mathcal{M}_t = \{(p_i^{KF}, p_i^t) \mid \|d_i^{KF} - d_i^t\|_2 \leq \epsilon\} \quad (11)$$

where $d_i^{KF}$ and $d_i^t$ are the descriptors of the $i$-th keypoint in the keyframe and the current frame, respectively, and $\epsilon$ is a threshold that defines the maximum allowable difference between the descriptors for a valid match. The set of valid correspondences is $\mathcal{M}_t$, where each pair ($p_i^{KF}, p_i^t$) represents a matched keypoint in the keyframe and the current frame.

### 3.1.4 Keyframe Selection

Keyframe selection plays a crucial role in maintaining the efficiency and robustness of visual SLAM. A frame $I_t$ is designated as a keyframe $KF_t$ if it satisfies a set of well-defined and rigorous selection criteria. The decision process can be expressed as:

$$KF_t = \begin{cases} I_t, & if f(I_t) = 1 \\ \emptyset, & otherwise \end{cases} \quad (12)$$

where $f(I_t)$ is a binary function that evaluates whether $I_t$ meets the keyframe selection conditions. The criteria for selecting a keyframe typically involve three primary conditions, which are critical for ensuring both the efficiency and accuracy of the SLAM system.

First, a keyframe is selected if the camera pose transformation between consecutive frames exceeds a predefined threshold $\tau_T$, ensuring sufficient motion variation:

$$\|T_t^{-1}T_{t-1} - I\|_F > \tau_T \quad (13)$$

where $T_t \in SE(3)$ represents the camera pose at time $t$, $I$ is the identity transformation matrix, and $\|\cdot\|_F$ denotes the Frobenius norm.

Second, the parallax of tracked features should be above a minimum threshold $\tau_p$, ensuring a significant viewpoint change to improve map accuracy:

$$\frac{1}{|\mathcal{M}_t|} \sum_{p_i \in \mathcal{M}_t} \|p_i^{t-1} - p_i^t\| > \tau_p \quad (14)$$

where $\mathcal{M}_t$ denotes the set of successfully tracked features between frames $I_{t-1}$ and $I_t$, and $p_i^t$ represents the 2D image coordinates of the $i$-th feature in frame $I_t$.

Third, a new keyframe is required when the number of successfully tracked features falls below a threshold $N_{min}$, indicating inadequate feature retention:

$$|\mathcal{M}_t| < N_{min} \quad (15)$$

where $|\mathcal{M}_t|$ is the total number of feature correspondences. If a frame is selected as a keyframe and feature descriptors have not yet been extracted, ORB feature extraction is performed to ensure distinctiveness for future tracking, loop closure detection, and map optimization. This process enhances system robustness by maintaining a well-distributed set of reliable keypoints.

### 3.2 Feature-based Method

The feature-based method is widely adopted in visual odometry due to its robustness against camera motion, lighting variations, and dynamic objects. This method follows a three-step process: first, it detects corner points in the original image; second, it extracts descriptors for these points; and finally, it matches corresponding points across different images based on descriptor similarity. Common feature detectors include SIFT [33], SURF [34], and ORB [35]. While SIFT and SURF exhibit strong rotational and scale invariance, their computational cost is relatively high. In contrast, ORB achieves significantly higher extraction speed, approximately 15 times faster than SURF, while preserving robustness to rotation and scale variations. This efficiency makes ORB particularly suitable for real-time applications.

The ORB detection process begins with filtering the original image, followed by scanning for FAST corners. Each detected corner is then described using the binary BRIEF descriptor. Feature matching is performed based on Hamming distance, where a smaller distance indicates a higher similarity between feature points. Although these steps theoretically suffice for camera pose estimation and visual odometry, real-world environments often introduce challenges such as repetitive textures, noise, or abrupt lighting changes. To enhance accuracy, additional filtering is required to remove mismatches and outliers.

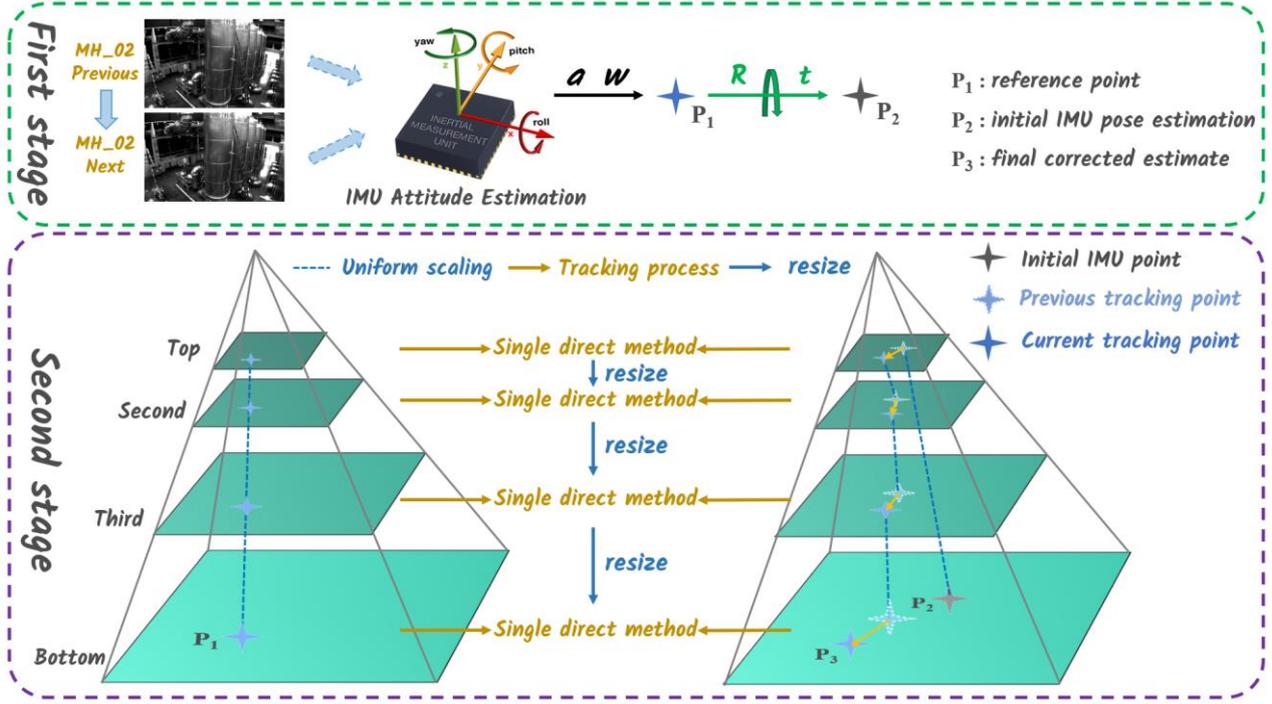

**Fig. 5** Two-Stage Fusion Method: Schematic Diagram of Stage I and Stage II

### 3.3 Direct Method

The optical flow method eliminates the need for descriptor extraction in corner matching by leveraging the strong assumption of grayscale consistency, significantly reducing computational overhead. Furthermore, the direct method builds upon optical flow, simultaneously performing both feature point matching and camera pose estimation to minimize global error. In the direct method, the spatial coordinates of feature points are initially determined using triangulation, with their positions expressed in the first frame's camera coordinate system. An initial estimate of the rotation matrix $R$ and translation vector $t$ between consecutive frames is obtained from the IMU. A nonlinear optimization algorithm is then applied to refine the estimated feature point coordinates in the second frame, along with more precise values for $R$ and $t$.

Unlike feature-based methods, which perform feature matching and camera pose estimation separately, the direct method inherently integrates these processes. As a result, errors propagate multiplicatively throughout the estimation process. The first-order derivative of the optimization objective with respect to the optimization variables, represented by the Jacobian matrix, is determined by the image intensity gradient. The magnitude of this gradient directly influences optimization performance. When the local image gradient is low or noise levels are high, the gradient information may become unreliable, potentially causing instability in the optimization process or even rendering the method ineffective.

### 3.4 Fusion Method

Building on the direct method, we propose a novel fusion approach to enhance the real-time performance of the algorithm. This method avoids extracting certain non-essential descriptors through hybrid feature-direct fusion strategy, which reduces the dependence on descriptors and realizes a lighter and faster visual inertial SLAM algorithm. As shown in Fig. 5, the first stage utilizes the IMU-predicted last frame bit-pose $T_{t-1 \to t}^{IMU}$ for coarse matching:

$$\hat{T}_t = T_{t-1} \cdot T_{t-1 \to t}^{IMU} \quad (16)$$

where $\hat{T}_t$ denotes the estimation of the current pose $T_t$, $T_{t-1}$ denotes the pose of the previous frame, and $T_{t-1 \to t}^{IMU}$ is the relative pose between neighboring frames predicted by IMU. The reference point $p_i^{t-1}$ in the previous frame is mapped to the initial estimation point $p_i^t$ in the current frame by the pose solving of IMU. The second stage minimizes the photometric error to optimize the matching point:

$$\mathcal{M}_t = \left\{ (p_i^{t-1}, p_i^t) \mid \arg \min_{T_t} \sum_i \|\varepsilon\|^2 \right\} \quad (17)$$

$$\varepsilon = I_{t-1}(p_i^{t-1}) - I_t(T_t \cdot p_i^t) \quad (18)$$

where $I_{t-1}$ and $I_t$ denote the image intensity values of the previous and current frame, respectively. The term $\varepsilon$ represents the photometric error between corresponding points in consecutive frames.

For further optimization, a four-layer equally scaled image pyramid is constructed. First, the initial estimation point $p_i^t$ is mapped to the top layer of the pyramid and the optimal estimation point in the top layer is obtained by direct method optimization. The point is then successively mapped down through each layer, undergoing direct method optimization at each stage until refinement is completed at the bottom layer. The final point $p_i^{t*}$ is the final estimation point after optimization. Through this optimization process from coarse to fine, the accurate bit-position estimation $T_t$ between two frames is obtained. Constructing a multi-layer image pyramid mitigates the non-convexity issues often encountered in direct method optimization, which can otherwise hinder convergence. Additionally, as image resolution decreases at higher layers, large estimation deviations in the bottom layer are transformed into smaller deviations at the top layer. This transformation reduces the sensitivity of the direct method to image motion, enhancing the algorithm's robustness.

## 4 Experimental Results

### 4.1 Experimental Setup and Evaluation

The EuRoC MAV dataset [36] is a widely used visual-inertial benchmark designed for indoor environments, specifically targeting the evaluation of algorithms for Micro Aerial Vehicles (MAVs). It encompasses two representative scenarios: an industrial plant and a standard room, each presenting unique challenges for visual-inertial systems. The dataset provides synchronized stereo images captured by high-resolution binocular cameras, as well as IMU data, including acceleration and angular velocity measurements recorded at 200 Hz. Additionally, the dataset includes high-precision reference trajectories, obtained using motion capture systems or laser trackers, serving as ground truth for performance evaluation. The industrial plant environment is characterized by a complex, cluttered layout with dynamic lighting conditions and various obstacles, making it suitable for testing the robustness of algorithms in real-world conditions. In contrast, the standard room environment offers a simpler, more controlled setting for evaluating system performance under less demanding conditions. By providing these diverse scenarios, the EuRoC MAV dataset serves as an essential benchmark for assessing the accuracy, robustness, and scalability of visual-inertial odometry and SLAM systems.

To assess the localization accuracy and computational efficiency of the HS-SLAM system, we conduct comparative experiments using the EuRoC MAV dataset and benchmark its performance against the state-of-the-art (SOTA) ORB-SLAM3 algorithm. The experiments are carried out on a laptop running Ubuntu 18.04 LTS, equipped with an Intel Core i5-8300H CPU, an NVIDIA GTX 1660 Ti GPU (6GB VRAM), and 8GB of RAM. To ensure fair comparison, we evaluate trajectory accuracy using three widely recognized metrics in the field: Absolute Trajectory Error (ATE), Relative Pose Error (RPE) and Standard Deviation (S.D.).

For a rigorous and consistent performance evaluation, we employ the EVO evaluation tool, a widely adopted framework for assessing visual odometry and SLAM systems. EVO computes key metrics such as ATE, RPE, and S.D., which are essential for measuring the accuracy, consistency, and robustness of the estimated trajectories. The following equations are used to calculate each metric:

The ATE quantifies the global deviation between the estimated trajectory and the ground truth over the entire trajectory. It is computed as:

$$ATE_{all} = \sqrt{\frac{1}{N}\sum_{i=1}^{N} \left\| log\left(T_{gt,i}^{-1}T_{esti,i}\right)^{\vee} \right\|_2^2} \quad (19)$$

where $\mathbf{T}_{gt,i}$ and $\mathbf{T}_{esti,i}$ denote the ground truth and estimated transformation matrices at the $i$-th time step respectively, $N$ is the total number of poses, $\log(\cdot)$ maps the transformation to the Lie algebra $\mathfrak{se}(3)$, and $(\cdot)^{\vee}$ extracts the corresponding 6D error vector. The Euclidean norm $\|\cdot\|_2$ is used to compute the error magnitude.

The RPE evaluates the local consistency of the trajectory by comparing the relative motion between pairs of poses. It is defined as:

$$RPE_{all} = \sqrt{\frac{1}{N-\Delta t}\sum_{i=1}^{N-\Delta t} \left\| log\left(\left(T_{gt,i}^{-1}T_{gt,i+\Delta t}\right)^{-1}\left(T_{esti,i}^{-1}T_{esti,i+\Delta t}\right)\right)^{\vee} \right\|_2^2} \quad (20)$$

where $\Delta t$ is the time interval between compared poses, and all other terms are as defined above. This metric captures both rotational and translational inconsistencies in local motion.

Finally, the S.D. is computed to evaluate the variability of the trajectory error across all poses as:

$$\text{S.D.} = \frac{1}{N}\sum_{i=1}^{N} \left(\left\| log\left(T_{gt,i}^{-1}T_{esti,i}\right)^{\vee} \right\|_2 - \mu\right)^2 \quad (21)$$

where $\mu$ is the mean of the ATE values over the entire trajectory.

Using the EVO tool, we generate error plots and performance metrics, providing a comprehensive analysis of trajectory accuracy, system robustness, and consistency under various conditions.

### 4.2 Localization Accuracy

For localization accuracy assessment, the HS-SLAM algorithm was evaluated on five dataset sequences: MH01~MH04 and V102, with trajectory lengths ranging from 73.473 m to 130.928 m. Using the EVO evaluation tool, we systematically quantified the deviation of the estimated trajectories from the ground truth and compared HS-SLAM against the state-of-the-art ORB-SLAM3 algorithm in terms of ATE and RPE.

As shown in Fig. 6, the second row of trajectory plots represents the ground truth (black dashed lines), the ORB-SLAM3 estimates (blue solid lines), and the HS-SLAM estimates (green solid lines). The first row contains zoomed-in views of selected regions for closer inspection, with purple dashed lines indicating the corresponding areas. Overall, the three trajectories are closely aligned, particularly in the MH04 sequence. This alignment is attributed to both HS-SLAM and the state-of-the-art ORB-SLAM3 algorithm achieving centimeter-level accuracy. The trajectory lengths in the dataset, however, are on the order of meters, which results in a two-order-of-magnitude difference between the algorithms' performance. In the zoomed-in view of the MH02 sequence, the HS-SLAM trajectory (green) is noticeably closer to the ground truth than that of ORB-SLAM3 (blue). This observation demonstrates that HS-SLAM achieves higher accuracy compared to ORB-SLAM3. The bottom row shows the projection of trajectory errors onto the x, y, and z axes. All three trajectories are nearly indistinguishable. This qualitative analysis suggests that both algorithms exhibit comparable localization accuracy, with HS-SLAM showing a slight advantage.

To further highlight the performance differences between the two algorithms, which cannot be fully captured through visual inspection, we present heatmaps for additional analysis.

Fig. 7. presents a heatmap comparison of absolute trajectory errors between the two algorithms, where error magnitudes are visually represented by color. Red indicates larger errors, while blue represents smaller errors. The bars on the right side of the heatmap display the maximum, median, and minimum errors.

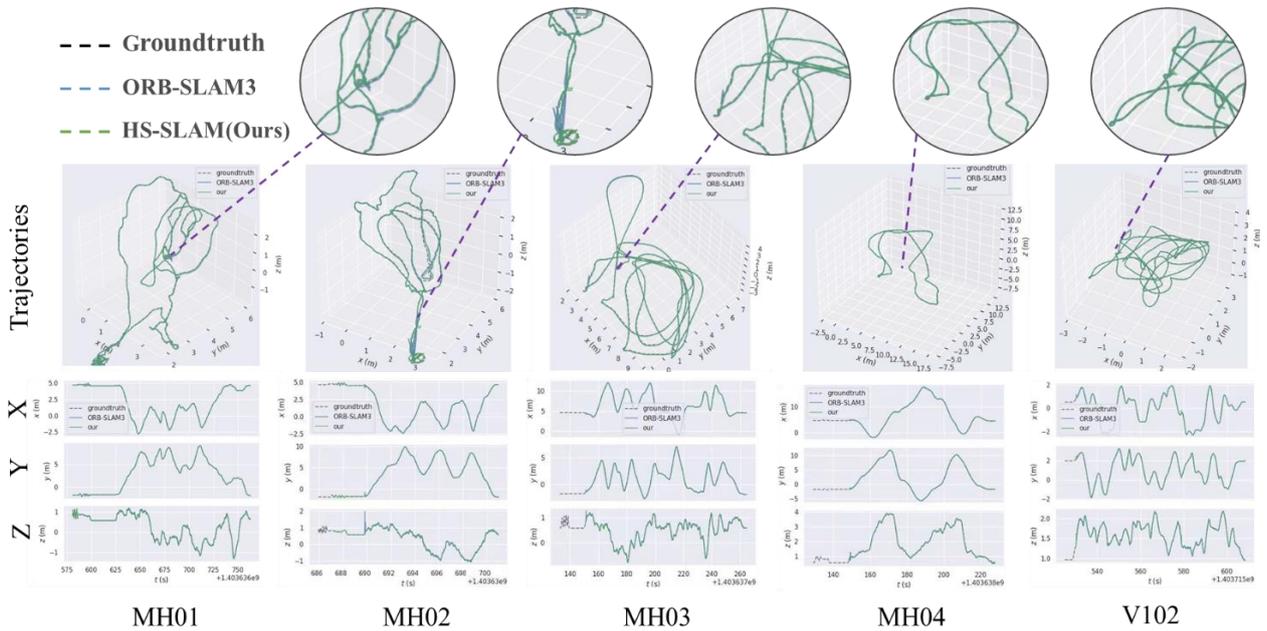

**Fig. 6** Performance evaluation of HS-SLAM and ORB-SLAM3: Trajectory visualization and X/Y/Z-axis error assessment on five EuRoC sequences

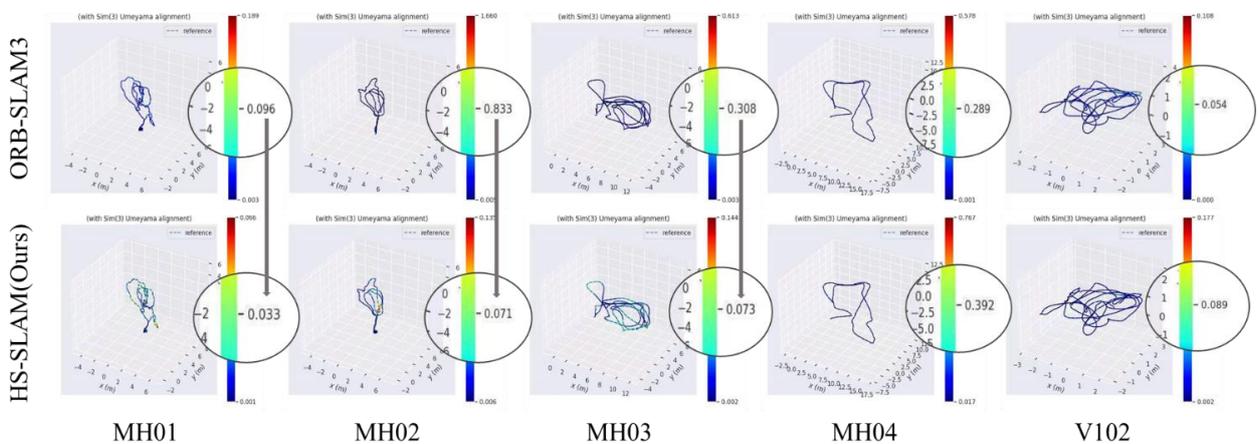

**Fig. 7** ATE heatmap comparison in translation between HS-SLAM and ORB-SLAM3 on EuRoC sequences

Table 2. Performance comparison in the EuRoC dataset (RMSE ATE and S.D. in m, scale error in %).

| Sequences | ORB-SLAM3 | | HS-SLAM(Ours) | | Improvements | |
|---|---|---|---|---|---|---|
| | RMSE | S.D. | RMSE | S.D. | RMSE | S.D. |
| MH01 | 0.0259 | 0.0115 | 0.0209 | 0.0101 | 19.31% | 12.17% |
| MH02 | 0.0758 | 0.0624 | 0.0396 | 0.0238 | **47.76%** | **61.86%** |
| MH03 | 0.0437 | 0.0281 | 0.0367 | 0.0178 | 16.12% | 36.65% |
| MH04 | 0.0661 | 0.0348 | 0.0694 | 0.0274 | -5.00% | 21.26% |
| V102 | 0.0153 | 0.0085 | 0.0198 | 0.0117 | -29.41% | -37.65% |

Table 3. Performance comparison in the EuRoC dataset (RMSE RPE and S.D. in m, scale error in %).

| Sequences | ORB-SLAM3 | | HS-SLAM(Ours) | | Improvements | |
|---|---|---|---|---|---|---|
| | RMSE | S.D. | RMSE | S.D. | RMSE | S.D. |
| MH01 | 0.0090 | 0.0083 | 0.0060 | 0.0051 | 33.33% | 12.17% |
| MH02 | 0.0537 | 0.0533 | 0.0396 | 0.0057 | **47.76%** | 38.55% |
| MH03 | 0.0193 | 0.0185 | 0.0124 | 0.0107 | 35.75% | **42.16%** |
| MH04 | 0.0193 | 0.0185 | 0.0167 | 0.0142 | 13.47% | 23.24% |
| V102 | 0.0079 | 0.0067 | 0.0134 | 0.0116 | **-69.62%** | **-73.13%** |

Table 4. Image tracking time comparison (total frames, total tracking time in s, average tracking time per frame in ms).

| Sequences | Total Frames | Total Tracking Time | | Average Tracking Time per Frame | | Improvements |
|---|---|---|---|---|---|---|
| | | ORB-SLAM3 | HS-SLAM(Ours) | ORB-SLAM3 | HS-SLAM(Ours) | |
| MH01 | 3682 | 87.53 | 63.10 | 23.77 | 17.14 | **27.89%** |
| MH02 | 3040 | 72.86 | 50.96 | 23.97 | 16.76 | **30.08%** |
| MH03 | 2703 | 61.83 | 57.81 | 22.87 | 21.39 | 6.47% |
| MH04 | 2033 | 43.16 | 39.03 | 21.23 | 19.20 | 9.6% |
| V102 | 1713 | 35.21 | 34.65 | 20.55 | 20.23 | 1.59% |

As for the median error values, it is clear that HS-SLAM demonstrates significantly better performance than ORB-SLAM3 in the MH01 to MH03 sequences. The median errors for HS-SLAM in these sequences are consistently lower, indicating a more accurate trajectory estimate. However, in the MH04 and V102 sequences, ORB-SLAM3 has a slight advantage, with lower median errors compared to HS-SLAM. This analysis demonstrates that while HS-SLAM excels in most sequences, ORB-SLAM3 performs better in the MH04 and V102 sequences.

For a more precise quantitative measurement, we repeated the experiments five times and computed the ATE and RPE of the translational component, as shown in Table 2 and Table 3. HS-SLAM demonstrates superior performance on the MH01~MH04 sequences, with estimated trajectories closely aligning with the ground truth. In particular, HS-SLAM excels in the MH02 and MH03 sequences, with notable improvements of 47.76% and 16.12% in RMSE ATE, as well as a substantial reduction in standard deviation (S.D.).

However, its performance on the V102 sequence is weaker, with -29.41% worse RMSE ATE and -73.13% decrease in S.D.. This discrepancy arises because ORB-SLAM3 performs exceptionally well on the V102 sequence, achieving millimeter-level accuracy, causing HS-SLAM's deviations to appear larger in percentage terms. Despite this, HS-SLAM still maintains good accuracy with RMSE ATE of 0.0198 m in MH04 and 0.0134 m in V102.

Considering the experimental results, HS-SLAM demonstrates superior performance in four sequences and performs worse in one. This suggests that HS-SLAM offers a slight edge in localization accuracy compared to ORB-SLAM3 across the majority of sequences.

### 4.3 Computational Efficiency Assessment

We evaluated the computational performance by comparing the per-frame processing times of ORB-SLAM3 and the proposed HS-SLAM algorithm. As shown in Table 4, HS-SLAM significantly improves processing efficiency over the baseline method. The results indicate that ORB-SLAM3 achieves an image tracking frame rate of 40 to 50 fps, while HS-SLAM delivers a higher frame rate of 50 to 65 fps across the same test sequences, demonstrating a notable performance boost.

Specifically, HS-SLAM reduces the average processing time per frame by approximately 30% in the first two test sequences (MH01 and MH02) compared to ORB-SLAM3, with improvements of 27.89% and 30.08%, respectively. These results highlight the algorithm's efficiency, achieving both higher frame rates and a significant reduction in per-frame processing time. Across all five test sequences, HS-SLAM achieves an overall average speedup of 15.13%.

## 5 Conclusion

We propose HS-SLAM, a lightweight and fast visual-inertial SLAM algorithm designed to enhance localization efficiency. HS-SLAM employs a strategy that eliminates feature point and descriptor extraction for non-critical frames, resulting in an average tracking time improvement of 15% while maintaining slightly higher localization accuracy than ORB-SLAM3. This improvement is particularly beneficial for small embedded devices. The direct method assumes gray-scale invariance, which may fail under abrupt illumination changes. Future SLAM algorithms for mobile and wearable devices must adapt to complex outdoor conditions. While single-theory approaches excel in specific cases, robust solutions for diverse environments require further research.

## Compliance with Ethical Standards

**Conflict of interest** On behalf of all the authors, the corresponding author states that there is no conflict of interest.

## References

1. S. Thrun, W. Burgard, and D. Fox, Probabilistic Robotics (Intelligent Robotics and Autonomous Agents). Cambridge, MA, USA: MIT Press, 2005.
2. H. Durrant-Whyte and T. Bailey, "Simultaneous localization and mapping: Part I," IEEE Robot. Autom. Mag., vol. 13, no. 2, pp. 99–110, Jun. 2006.
3. J. Cheng, L. Zhang, Q. Chen, X. Hu, and J. Cai, "A review of visual SLAM methods for autonomous driving vehicles," Eng. Appl. Artif. Intell., vol. 114, Sep. 2022, Art. no. 104992.
4. Z. Huai and G. Huang, "A consistent parallel estimation framework for visual-inertial SLAM," IEEE Trans. Robot., vol. 40, pp. 3734–3755, Jul. 2024.
5. X. Liu, S. Wen, J. Zhao, T. Z. Qiu, and H. Zhang, "Edge-Assisted Multi-Robot Visual-Inertial SLAM With Efficient Communication," IEEE Trans. Autom. Sci. Eng., vol. 22, pp. 2186–2198, 2025.
6. Y. Ge, L. Zhang, Y. Wu, and D. Hu, "PIPO-SLAM: Lightweight Visual-Inertial SLAM With Preintegration Merging Theory and Pose-Only Descriptions of Multiple View Geometry," IEEE Trans. Robotics, vol. 40, pp. 2046–2059, Feb. 2024.
7. C. Liu, H. Yu, P. Cheng, W. Sun, J. Civera, and X. Chen, "PE-VINS: Accurate Monocular Visual-Inertial SLAM With Point-Edge Features," IEEE Trans. Intell. Vehicles, pp. 1–11, Jun. 2024.
8. I. A. Kazerouni, L. Fitzgerald, G. Dooly, and D. Toal, "A survey of state-of-the-art on visual SLAM," Expert Syst. Appl., vol. 205, p. 117734, Nov. 2022.
9. Z. Fan, L. Zhang, X. Wang, Y. Shen, and F. Deng, "LiDAR, IMU, and camera fusion for simultaneous localization and mapping: a systematic review," Artif. Intell. Rev., vol. 58, no. 6, p. 174, Mar. 2025.
10. R. Mur-Artal and J. D. Tardós, "ORB-SLAM2: An open-source SLAM system for monocular, stereo, and RGB-D cameras," IEEE Trans. Robot., vol. 33, no. 5, pp. 1255–1262, Oct. 2017.
11. P. van Goor, T. Hamel, and R. Mahony, "Equivariant filter (EqF)," IEEE Trans. Autom. Control, vol. 68, no. 6, pp. 3501–3512, Jun. 2023.
12. P. van Goor and R. Mahony, "EqVIO: An equivariant filter for visual-inertial odometry," IEEE Trans. Robot., vol. 39, no. 5, pp. 3567–3585, Oct. 2023.
13. C. Chen, P. Geneva, Y. Peng, W. Lee, and G. Huang, "Monocular Visual-Inertial Odometry with Planar Regularities," in 2023 IEEE Int. Conf. Robotics Autom. (ICRA), London, United Kingdom, 2023, pp. 6224–6231.
14. T. Qin, P. Li, and S. Shen, "VINS-Mono: A Robust and Versatile Monocular Visual-Inertial State Estimator," IEEE Trans. Robotics, vol. 34, no. 4, pp. 1004–1020, Aug. 2018.
15. C. Campos, R. Elvira, J. J. G. Rodríguez, J. M. M. Montiel, and J. D. Tardós, "ORB-SLAM3: An Accurate Open-Source Library for Visual, Visual–Inertial, and Multimap SLAM," IEEE Trans. Robotics, vol. 37, no. 6, pp. 1874–1890, Dec. 2021.
16. L. v. Stumberg and D. Cremers, "DM-VIO: Delayed Marginalization Visual-Inertial Odometry," IEEE Robotics Autom. Lett., vol. 7, no. 2, pp. 1408–1415, Apr. 2022.
17. X. Cai, L. Zhang, C. Li, G. Li, and T. H. Li, "VONAS: Network Design in Visual Odometry using Neural Architecture Search," in Proc. 28th ACM Int. Conf. Multimedia (MM '20), Seattle, WA, USA, 2020, pp. 727–735.
18. L. Liu, G. Li, and T. H. Li, "ATVIO: Attention Guided Visual-Inertial Odometry," in 2021 IEEE Int. Conf. Acoustics, Speech and Signal Processing (ICASSP), Toronto, ON, Canada, 2021, pp. 4125–4129.
19. A. J. Davison, I. D. Reid, N. D. Molton, and O. Stasse, "MonoSLAM: Real-Time Single Camera SLAM," IEEE Trans. Pattern Anal. Mach. Intell., vol. 29, no. 6, pp. 1052–1067, Jun. 2007.
20. R. Mur-Artal, J. M. M. Montiel, and J. D. Tardós, "ORB-SLAM: A Versatile and Accurate Monocular SLAM System," IEEE Trans. Robotics, vol. 31, no. 5, pp. 1147–1163, Oct. 2015.
21. H. Yang, J. Yuan, Y. Gao, X. Sun, and X. Zhang, "UPLP-SLAM: Unified point-line-plane feature fusion for RGB-D visual SLAM," Inf. Fusion, vol. 96, pp. 51–65, 2023.
22. X. Jia, Y. Ning, D. Chai, J. Fan, Z. Yang, X. Xi, F. Zhu, and W. Wang, "EGLT-SLAM: Real-Time Visual-Inertial SLAM Based on Entropy-Guided Line Tracking," IEEE Sens. J., vol. 24, no. 20, pp. 32757–32771, Oct. 2024.
23. H. M. S. Bruno and E. L. Colombini, "LIFT-SLAM: A deep-learning feature-based monocular visual SLAM method," Neurocomputing, vol. 455, pp. 97–110, 2021.
24. C. Shao, L. Zhang, and W. Pan, "Faster R-CNN Learning-Based Semantic Filter for Geometry

Estimation and Its Application in vSLAM Systems," IEEE Trans. Intell. Transp. Syst., vol. 23, no. 6, pp. 5257–5266, Jun. 2022.
25. R. A. Newcombe, S. J. Lovegrove, and A. J. Davison, "DTAM: Dense tracking and mapping in real-time," in 2011 Int. Conf. Computer Vision (ICCV), Barcelona, Spain, 2011, pp. 2320–2327.
26. J. Engel, T. Schöps, and D. Cremers, "LSD-SLAM: Large-Scale Direct Monocular SLAM," in Computer Vision – ECCV 2014, D. Fleet, T. Pajdla, B. Schiele, and T. Tuytelaars, Eds., Cham, Switzerland: Springer, 2014, pp. 834–849.
27. J. Engel, V. Koltun, and D. Cremers, "Direct Sparse Odometry," IEEE Trans. Pattern Anal. Mach. Intell., vol. 40, no. 3, pp. 611–625, Mar. 2018.
28. C. Forster, M. Pizzoli, and D. Scaramuzza, "SVO: Fast semi-direct monocular visual odometry," in 2014 IEEE Int. Conf. Robotics Autom. (ICRA), Hong Kong, China, 2014, pp. 15–22.
29. S. Leutenegger, S. Lynen, M. Bosse, R. Siegwart, and P. Furgale, "Keyframe-based visual–inertial odometry using nonlinear optimization," Int. J. Robotics Res., vol. 34, no. 3, pp. 314–334, 2014.
30. S. Tsuei, S. Soatto, P. Tabuada, and M. B. Milam, "Learned Uncertainty Calibration for Visual Inertial Localization," in 2021 IEEE Int. Conf. Robotics Autom. (ICRA), Xi'an, China, 2021, pp. 5311–5317.
31. H. Yin, S. Li, Y. Tao, J. Guo, and B. Huang, "Dynam-SLAM: An Accurate, Robust Stereo Visual-Inertial SLAM Method in Dynamic Environments," IEEE Trans. Robotics, vol. 39, no. 1, pp. 289–308, Feb. 2023.
32. J. Yin, A. Li, W. Xi, W. Yu, and D. Zou, "Ground-Fusion: A Low-cost Ground SLAM System Robust to Corner Cases," in 2024 IEEE Int. Conf. Robotics Autom. (ICRA), Yokohama, Japan, 2024, pp. 8603–8609.
33. D. G. Lowe, "Distinctive Image Features from Scale-Invariant Keypoints," Int. J. Computer Vision, vol. 60, no. 2, pp. 91–110, Nov. 2004.
34. H. Bay, A. Ess, T. Tuytelaars, and L. Van Gool, "Speeded-Up Robust Features (SURF)," Computer Vision and Image Understanding, vol. 110, no. 3, pp. 346–359, 2008.
35. E. Rublee, V. Rabaud, K. Konolige, and G. Bradski, "ORB: An efficient alternative to SIFT or SURF," in 2011 Int. Conf. Comput. Vis. (ICCV), Barcelona, Spain, 2011, pp. 2564–2571.
36. Burri M, Nikolic J, Gohl P, et al. The EuRoC micro aerial vehicle datasets. Int J Robot Res. vol. 35, no. 10, pp. 1157–1163, 2016.